\pdfoutput=1

\documentclass[11pt]{article}

\usepackage{amsmath,amsfonts,bm}

\def\eqref#1{equation~\ref{#1}}

\def\1{\bm{1}}

\DeclareMathAlphabet{\mathsfit}{\encodingdefault}{\sfdefault}{m}{sl}
\SetMathAlphabet{\mathsfit}{bold}{\encodingdefault}{\sfdefault}{bx}{n}

\DeclareMathOperator*{\argmax}{arg\,max}

\usepackage[final]{acl}

\usepackage{times}
\usepackage{latexsym}

\usepackage[T1]{fontenc}

\usepackage{amsmath,amssymb,amsfonts,amsthm,mathtools}
\usepackage{bm,bbm}
\usepackage{duckuments}
\usepackage{enumitem}
\usepackage{graphicx}
\usepackage{float}
\usepackage[multiple]{footmisc}
\usepackage{caption,subcaption}
\usepackage{cprotect}  %
\usepackage{textcomp}
\usepackage{stfloats}
\usepackage{hyperref}
\usepackage[capitalize,noabbrev]{cleveref}
\usepackage{listings}
\usepackage{lipsum}
\usepackage{longtable,tabularx,booktabs,wrapfig}
\usepackage{multirow}
\usepackage{siunitx}
\usepackage{url}
\usepackage{xcolor}
\usepackage{xspace}
\usepackage[utf8]{inputenc}
\usepackage{tcolorbox,tabularray}
\usepackage{placeins}
\usepackage{colortbl}

\usepackage{microtype}

\usepackage{inconsolata}

\definecolor{cornellred}{rgb}{0.7, 0.11, 0.11}

\newcommand{\sname}{PriME}

\definecolor{Gray2}{gray}{0.5}
\definecolor{RowHighlight}{gray}{0.9}

\theoremstyle{plain}

\theoremstyle{definition}

\theoremstyle{remark}

\usepackage[textsize=tiny]{todonotes}

\title{Personalized Language Models via\\Privacy-Preserving Evolutionary Model Merging}

\author{
 \textbf{Kyuyoung Kim\textsuperscript{1}},
 \textbf{Jinwoo Shin\textsuperscript{1}},
 \textbf{Jaehyung Kim\textsuperscript{2}}
\\
 \textsuperscript{1}KAIST AI,
 \textsuperscript{2}Yonsei University
\\
  \texttt{\{kykim,jinwoos\}@kaist.ac.kr}, \texttt{jaehyungk@yonsei.ac.kr}
}

\begin{document}
\maketitle

\begin{abstract}
Personalization in language models aims to tailor model behavior to individual users or user groups.
Prompt-based methods incorporate user preferences into queries, while training-based methods encode them into model parameters.
Model merging has also been explored for personalization under limited data.
However, existing methods often fail to directly optimize task-specific utility and lack explicit mechanisms for privacy preservation.
To address the limitations, we propose Privacy-Preserving Model Merging via Evolutionary Algorithms ({\sname}), a novel personalization approach that employs gradient-free methods to directly optimize utility while reducing privacy risks.
By integrating privacy preservation into the optimization objective, {\sname} creates personalized modules that effectively capture target user preferences while minimizing privacy risks for data-sharing users.
Experiments on the LaMP benchmark show that {\sname} consistently outperforms a range of baselines, achieving up to a 45\% improvement in task performance.
Further analysis demonstrates that {\sname} achieves a superior privacy-utility trade-off compared to a prior state-of-the-art, with enhanced robustness to membership inference attacks and greater utility in capturing user preferences.
\end{abstract}

\section{Introduction}
\label{s:intro}

Pre-trained on web-scale data, large language models (LLMs) have emerged as powerful tools~\citep{radford2018improving,brown2020language}, capable of performing a wide range of natural language processing (NLP) tasks, from conversational agents, translation, and question-answering to code generation~\citep{achiam2023gpt,team2024codegemma}.
Notably, LLMs have demonstrated remarkable effectiveness as generalist models, able to perform complex tasks with little to no task-specific data~\citep{radford2019language}.
To extend their utility beyond general-purpose applications, LLM personalization seeks to tailor model responses to the preferences of individual users or user groups~\citep{salemi2024lamp}.
Personalized LLMs are particularly important in domains such as education~\citep{kasneci2023chatgpt}, healthcare~\citep{liu2023large}, and content recommendation~\citep{baek2024knowledge} to enhance performance and user experience.

Previous approaches to LLM personalization can be categorized as methods based on prompt augmentation, training, and, more recently, model merging.
Prompt-based methods incorporate user preferences or relevant historical data into input prompts, enabling models to better understand and respond to user-specific queries~\cite{mysore2024pearl,richardson2023integrating,kim2025few}.
While these methods are relatively simple to implement and avoid additional training, the performance is inherently constrained by the model's context length and may result in significantly higher inference costs due to enlarged input prompts.

In contrast, training-based approaches aim to capture user preferences directly in model parameters.
One straightforward method is to train a parameter-efficient fine-tuning (PEFT) module for each user on the user's private data~\citep{tan2024democratizing}.
However, when data is limited, these methods may struggle to achieve effective personalization.
Moreover, individually owned modules limits opportunities for collaborative benefits through shared adaptation.
To address this, several methods leverage data from multiple users or groups to model group-level preferences~\citep{li2024steerability,zhao2024group}.
However, direct access to raw data from multiple users or groups is rare in practice and raises substantial privacy concerns, especially in the absence of explicit safeguards.
Model merging has also been explored for personalization under limited data~\citep{tan2024personalized}, but existing approaches do not directly optimize for task-specific utility or privacy.

To address the challenges, we propose \textbf{Pri}vacy-Preserving \textbf{M}odel Merging via \textbf{E}volutionary Algorithms ({\sname}), a novel LLM personalization approach that leverages evolutionary methods to utilize knowledge from a community of data-sharing users while minimizing the risk of privacy leakage.
Specifically, given a set of users who have consented to share their PEFT modules, {\sname} employs evolutionary algorithms to find optimal weights for model merging~\citep{wortsman2022model}, creating a personalized module that captures the preferences of a target user.
As each shared module is trained on private data, the merged module poses a potential risk of privacy leakage for sharing users, e.g., through membership inference attacks (MIA), a widely used approach for assessing practical privacy leakage~\citep{shokri2017membership}.
To this end, we incorporate privacy preservation into the merging process by optimizing similarity with the target user while minimizing the average similarity with the sharing users.
The gradient-free nature of evolutionary algorithms enables direct optimization of non-differentiable metrics, such as ROUGE, used to measure the prediction similarity.
This allows us to achieve a superior privacy-utility trade-off between capturing the target user's preferences and minimizing privacy risks for sharing users.

\begin{figure*}[t]
    \small
    \centering
    \includegraphics[width=0.9\linewidth]{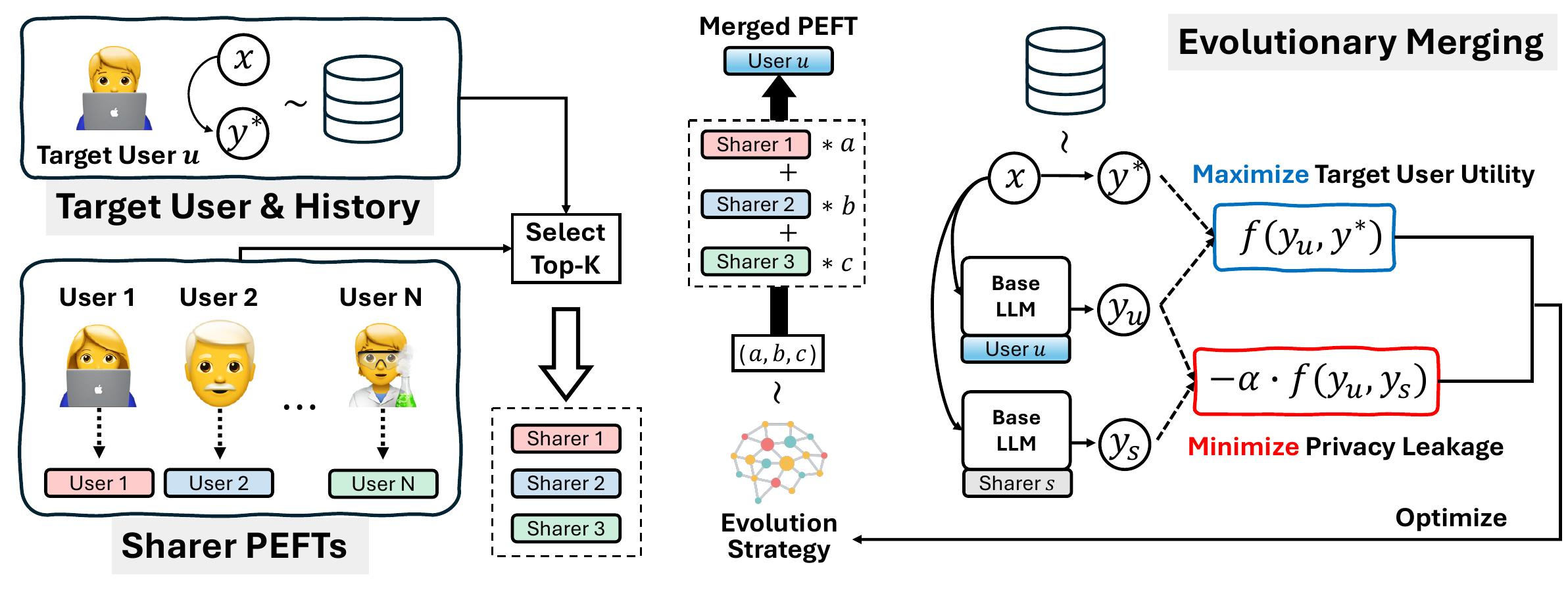}
    \vspace{-0.10in}
    \caption{\textbf{Overview of {\sname}.} {\sname} is an evolutionary method for LLM personalization that optimizes model merging to create a personalized, parameter-efficient fine-tuning module. Using gradient-free optimization, {\sname} directly optimizes utility metrics with a privacy-preserving mechanism to achieve a superior privacy-utility trade-off.}
    \label{fig:concept}
    \vspace{-0.15in}
\end{figure*}

Experimental results on LaMP~\citep{salemi2024lamp}, a widely used personalization benchmark consisting of various text classification and generation tasks, demonstrate that {\sname} consistently outperforms both prompt- and training-based baselines, achieving up to a 45\% improvement in performance.
Our findings show that {\sname} produces personalized modules that not only better capture user preferences but also exhibit greater robustness to MIA risks compared to a prior state-of-the-art merging method.
Qualitatively, we observe that with an appropriate choice of the parameter controlling the privacy-utility trade-off, the modules produced by {\sname} yield predictions that both better align with the target user's behavior patterns and are more distinct from those of the sharing users.
Notably, this is achieved without introducing additional complexity to the model architecture or training, unlike prior approaches~\citep{tan2024personalized}.
These results highlight the potential of evolutionary methods for achieving effective privacy-preserving LLM personalization.

Our main contributions are as follows:
\begin{itemize}
    \item We introduce {\sname}, an evolutionary approach that automates the discovery of optimal model merging recipes for privacy-preserving LLM personalization.
    \item To our knowledge, we are the first to examine privacy leakage via MIA in the context of merging-based LLM personalization, and to demonstrate that gradient-free approaches can effectively mitigate these risks.%
    \item We demonstrate that {\sname} outperforms both prompt- and training-based baselines on the LaMP benchmark, achieving superior privacy-utility trade-offs by producing personalized modules that better capture user preferences and exhibit enhanced robustness to MIA.%
\end{itemize}

\section{Preliminaries}
\label{s:prelim}

\subsection{LLM personalization}
Personalizing LLMs involves tailoring model responses to individual users or groups based on their historical behavior data.
Specifically, given an LLM $\mathcal{M}$ and history data $\mathcal{H}_u = \{ h_u \}$ for user $u$, we aim to generate an output $y$ that is closely aligned with the user's preferences conditioned on both the input $x$ and $\mathcal{H}_u$:
\begin{equation}
    \max_{y \sim \mathcal{M}(\cdot \mid x, \mathcal{H}_u)} f(y, y^{*}),
\end{equation}
where $f$ is a target utility metric, such as ROUGE, that measures the similarity between the generated response $y$ and the user-preferred ground-truth response $y^{*}$ to $x$.
Each history item $h_u \in \mathcal{H}_u$ may consist of either a $(x_u, y_u)$ pair in a task-specific query-response format, capturing how the user has historically responded to similar queries, or a text sequence that provides contextual information about the user's behavior patterns.

\subsection{Model merging for personalization}
Given a group of sharing users $\mathcal{S}$, each with a personalized PEFT $\Delta W_s$ trained on their private data, our goal is to create a module $\Delta W_u$ for a target user $u$ that aligns with the user's historical data $\mathcal{H}_u$ by leveraging the shared modules.
To achieve this, we (1) identify a subset of users $\bar{\mathcal{S}} \subset \mathcal{S}$ that are most relevant to the target user, along with the corresponding modules $\{ \Delta W_s \mid s \in \bar{\mathcal{S}} \}$, and (2) determine the optimal weights $\{ w_s \mid s \in \bar{\mathcal{S}} \}$ to linearly interpolate the selected modules.
Using the interpolation weights, we construct $\Delta W_u$ as
\begin{equation}
\scalebox{0.95}{%
$\begin{aligned}
    \Delta W_u = \sum_{s \in \bar{\mathcal{S}}} w_s \Delta W_s.
\end{aligned}$}
\end{equation}
If the weights are scalars, merging is performed at the module level, whereas if they are vectors, it occurs at a more granular level, such as per-layer LoRA weights.
Once we obtain a module $\Delta W_u$ for user $u$, we condition the LLM $\mathcal{M}$ on $\Delta W_u$ to generate personalized responses:
\begin{equation}
    \mathcal{M}_u(\cdot) = \mathcal{M}(\cdot \mid \Delta W_u),
\end{equation}
where $\mathcal{M}_u$ denotes the LLM with the personalized PEFT module $\Delta W_u$ integrated into $\mathcal{M}$.
For the remainder, we present our method using LoRA~\citep{hu2022lora} given its widespread adoption.

\subsection{Privacy risks in LLMs}
LLMs show strong performance across a wide range of tasks, but their reliance on large-scale data in various stages of training and inference introduces significant privacy risks.
For example, LLMs have been shown to memorize portions of their training data, often reproducing the data verbatim when prompted appropriately~\citep{carlini2023quantifying}.
Prior work has studied such privacy risks via methods such as MIAs in the contexts of pre-training~\citep{duan2024membership}, fine-tuning~\citep{mireshghallah2022empirical}, and prompt-based adaptation~\citep{duan2024privacy}.
As personalization involves sensitive user data, evaluating and mitigating such risks becomes even more crucial.
To our knowledge, we are the first to provide empirical assessments of privacy risks in merging-based personalization and to propose a method that explicitly addresses the challenges.

\section{Privacy-Preserving Evolutionary LLM Personalization}
\label{s:method}

In this section, we present {\sname}, a novel approach for achieving privacy-preserving LLM personalization via evolutionary model merging.

\subsection{Personalization via model merging}
\label{ss:prime}

\paragraph{User embedding and selection.}

Selecting only the most relevant LoRA modules for the target user is important for privacy, as it limits the number of modules trained on private data involved in merging.
To measure user similarity, we adopt a method from prior work~\citep{tan2024personalized}, where an encoder-only language model $\mathcal{E}$ encodes each history item $h_u \in \mathcal{H}_u$, and the user embedding is given by the average of the encoded items:
\begin{equation}
\scalebox{0.95}{%
$\begin{aligned}
    \mathbf{e}_u = \sum_{h_u \in \mathcal{H}_u} \mathcal{E}(h_u) / |\mathcal{H}_u|.   
\end{aligned}$}
\label{eq:user-embed}
\end{equation}
Given the user embedding $\mathbf{e}_u$ for the target user $u$ and $\mathbf{e}_s$ for each sharing user $s \in \mathcal{S}$, computed as in Eq.~\ref{eq:user-embed}, we compute the cosine similarity to select the top-$k$ most relevant sharing users:
\begin{equation}
\scalebox{0.82}{%
$\begin{aligned}
    \bar{\mathcal{S}} = \left\{ s \in \mathcal{S} \mid \text{cos}(\mathbf{e}_u, \mathbf{e}_s) \ge \text{top-$k$}_{s \in \mathcal{S}}(\text{cos}(\mathbf{e}_u, \mathbf{e}_s)) \right\}.
\end{aligned}$}
\label{eq:user-select}
\end{equation}
We empirically find that selecting only a few LoRA modules via this approach often suffices for effective personalization (see Section~\ref{ss:exp-main}).

\paragraph{Evolutionary merging for personalization.}

Inspired by the success of evolutionary approaches in developing strong foundation models via model merging~\citep{akiba2025evolutionary}, we extend the idea to automatically discover merging recipes for creating personalized LoRA modules.
Specifically, we employ an evolutionary algorithm, such as CMA-ES~\citep{hansen2006cma}, to find the optimal weights for linearly interpolating a subset of LoRAs $\{ \Delta W_s \mid s \in \bar{\mathcal{S}} \}$, selected according to Eq.~\ref{eq:user-select}.
The algorithm searches for optimal interpolation weights over a fixed number of iterations based on how well the resulting merged module aligns with the preferences of the target user $u$:
\begin{equation}
    \argmax_{\{ w_s \mid s \in \bar{\mathcal{S}} \}} \, \mathbb{E}_{y \sim \mathcal{M}_u} \left[ f(y, y^{*}) \right],
\end{equation}
where $\mathcal{M}$ is an LLM, and $f$ is a target utility metric that measures the similarity between the response $y$, generated by the LLM conditioned on $\Delta W_u$, and the user-preferred response $y^{*}$.
As evolutionary algorithms are gradient-free, one can directly optimize $f$, which is often non-differentiable.
Notably, $f$ can also be a suitable combination of multiple utility metrics relevant to the task.

\subsection{Threat model and membership inference}

Constructing a personalized module from a set of LoRAs raises concerns for the sharing users due to  potential privacy leakage, e.g., via membership inference attacks (MIA)~\citep{shokri2017membership,yeom2018privacy}.
Specifically, we consider an adversary with black-box access to the personalized LLM $\mathcal{M}_u$, conditioned on the module $\Delta W_u$, aiming to infer whether a particular data point is a member of the training data for a sharing user involved in the merging.
The adversary can query $\mathcal{M}_u$ with a set of task-specific input-response pairs $(x_u, y_u)$ and obtain the model's loss on each $y_u$.
Following prior MIA approaches, we use the loss as a score, which is thresholded to infer the membership of a data point~\cite{yeom2018privacy,duan2024membership}.

\subsection{Optimizing privacy preservation}
\label{ss:prime-privacy}

\paragraph{Privacy measure.}
To mitigate the MIA risks, we propose not only maximizing alignment with the target user's preferences but also minimizing similarity with the sharing users selected for merging.
Specifically, for a sharing user $s$, we measure this similarity using the utility metric $f$, where predictions from the sharing user's LoRA $\Delta W_s$ serve as reference labels and those from the merged LoRA $\Delta W_u$ as regular predictions.
Assuming access only to the target user's data, we compute the similarity over the target user's history $\mathcal{H}_u$ as follows:
\begin{equation}
\scalebox{0.95}{
$\begin{aligned}
    P_u(s) = \sum_{h_u \in \mathcal{H}_u} f(\mathcal{M}_u(h_u), \mathcal{M}_s(h_u)) / |\mathcal{H}_u|.
\label{eq:ppreserve}
\end{aligned}$}
\end{equation}
Intuitively, this evaluates the extent to which information about the sharing users can potentially be inferred from the merged LoRA $\Delta W_u$.

\paragraph{Privacy-aware optimization.}
LLM personalization can be viewed as a multi-objective optimization, where we maximize alignment with the target user's preferences while minimizing the privacy risks associated with the shared components.
To this end, we optimize the following objective, leveraging the privacy measure defined in Eq.~\ref{eq:ppreserve}:
\begin{equation}
\scalebox{0.95}{
$\begin{aligned}
    \argmax_{\{ w_s \mid s \in \bar{\mathcal{S}} \}}\, \mathbb{E}_{y \sim \mathcal{M}_u} \left[ f(y, y^{*}) \right] - \alpha \cdot \sum_{s \in \bar{\mathcal{S}}} \frac{P_u(s)}{|\bar{\mathcal{S}}|},
\label{eq:ppreserve-obj}
\end{aligned}$}
\end{equation}
where $\alpha$ controls the trade-off between capturing the target user's preferences and minimizing the similarity with the sharing users.
The use of $f$ in the privacy measure allows us to account for task-specific prediction similarities and to naturally define an objective that balances privacy and utility.
We empirically find that, for a suitable $\alpha$, the resulting module yields predictions that are more distinct from those of sharing users, with greater robustness to MIA compared to competing methods.
Notably, this is achieved while significantly better capturing the target user's preferences.

\begin{algorithm}[t]
  \caption{{\sname} algorithm}
  \label{alg:main}
\begin{algorithmic}
  \footnotesize  %
  \STATE
  \textbf{Input:} LLM $\mathcal{M}$, target user $u$, target user history $\mathcal{H}_u$, target user embedding $\mathbf{e}_u$, sharing users $\mathcal{S}$, sharing user embeddings $\{ \mathbf{e}_s \}$, shared LoRAs $\{ \Delta W_s \}$, target utility metric $f$, privacy coefficient $\alpha$, optimization budget $b$
  \vspace{0.05in} 
  \hrule
  \vspace{0.05in}
  \STATE \texttt{\color{Gray2} /* Select similar sharing users */}
  \STATE $\bar{\mathcal{S}} \leftarrow \left\{ s \in \mathcal{S} \mid \text{cos}(\mathbf{e}_u, \mathbf{e}_s) \ge \text{top-$k$}_{s \in \mathcal{S}}(\text{cos}(\mathbf{e}_u, \mathbf{e}_s)) \right\}$

  \STATE \texttt{\color{Gray2} /* Evolve interpolation weights */}
  \STATE $\texttt{solver} \leftarrow \texttt{EvolutionStrategy()}$
  \FOR{$t=0$ {\bfseries to} $b-1$}
  \STATE $\hat{w} \leftarrow \texttt{solver.ask}()$
  \STATE \texttt{\color{Gray2} /* Adapt with proposed weights */}
  \STATE $\Delta \hat{W_u} \leftarrow \sum_{s \in \bar{\mathcal{S}}} \hat{w}(s) \Delta W_s$
  \STATE $\hat{\mathcal{M}_u}(\cdot) \leftarrow \mathcal{M}(\cdot \mid \Delta \hat{W_u})$

  \STATE \texttt{\color{Gray2} /* Evaluate candidate $\hat{\mathcal{M}_u}$ */}
  \STATE $U_u \leftarrow \sum_{h_u \in \mathcal{H}_u} f(\hat{\mathcal{M}_u}(h_u), y^{*}(h_u)) / |\mathcal{H}_u|$
  \STATE $P_u \leftarrow \{ \sum_{h_u \in \mathcal{H}_u} f(\hat{\mathcal{M}_u}(h_u), \mathcal{M}_s(h_u)) / |\mathcal{H}_u| \}$
  \STATE $r \leftarrow U_u - \alpha \cdot \sum_{s \in \bar{\mathcal{S}}} P_u(s) / |\bar{\mathcal{S}}|$
  \STATE \texttt{\color{Gray2} /* Update optimal weights */}
  \STATE $\texttt{solver.tell(}r\texttt{)}$
  \STATE $w \leftarrow \texttt{solver.results()}$

  \ENDFOR
  \STATE \texttt{\color{Gray2} /* Create $\mathcal{M}_u$ with optimal weights */}
  \STATE $\Delta W_u \leftarrow \sum_{s \in \bar{\mathcal{S}}} w(s) \Delta W_s$
  \STATE $\mathcal{M}_u(\cdot) \leftarrow \mathcal{M}(\cdot \mid \Delta W_u)$
\STATE \textbf{return} personalized LLM $\mathcal{M}_u$ for user $u$
\end{algorithmic}
\end{algorithm}

\section{Experiments}
\label{s:experiments}

We evaluate {\sname} in terms of (a) its effectiveness in capturing target user preferences and (b) its ability to mitigate privacy risks for sharing users.

\subsection{Setup}

\paragraph{Benchmark and tasks.}

For our main evaluation, we use LaMP~\citep{salemi2024lamp}, a widely used benchmark of personalized text classification and generation tasks.
Specifically, we focus on three classification tasks (LaMP-1, LaMP-2, and LaMP-3) and three generation tasks (LaMP-4, LaMP-5, and LaMP-7).
We follow the approach from prior work for data splitting and selecting sharing users~\citep{tan2024personalized}.
See \cref{s:appendix:benchmark-details,s:appendix:exp-details} for details on the benchmark and setup.\footnote{Code and implementation details are available at \url{https://github.com/kykim0/PriME}}

\paragraph{Baselines.}
We evaluate {\sname} along with several state-of-the-art baselines.
For prompt-based methods, we consider retrieval-augmented generation (RAG)~\citep{salemi2024lamp}, which augments queries with retrieved history items, and profile-augmented generation (PAG)~\citep{richardson2023integrating}, which adds a textual summary of the user's profile to the prompt.
We also assess Per-Pcs~\citep{tan2024personalized}, a competing merging-based method that creates personalized modules by autoregressively merging relevant sharing modules, with selection guided by \textit{gating} vectors.
We use Llama-2-7b~\citep{touvron2023llama} as the base model and BM25~\citep{robertson2009probabilistic} for retrieval unless stated otherwise.

\begin{table*}[ht]
    \centering
    \scriptsize  %
    \setlength{\tabcolsep}{5pt}
    \begin{tabular}{llcccccccccc|ccc}
        \toprule[1pt]
        \multirow{2.5}{*}{\textbf{Task}} & \multirow{2.5}{*}{\textbf{Metric}} & \textbf{NP} & \textbf{RAG} & \multicolumn{2}{c}{\textbf{PAG}} & \multicolumn{3}{c}{\textbf{\textsc{Per-Pcs}}} & \multicolumn{3}{c}{\textbf{\textsc{\sname} (Ours)}} & \multicolumn{3}{c}{\textbf{OPPU}} \\ 
        \cmidrule(lr){3-3} \cmidrule(lr){4-4} \cmidrule(lr){5-6} \cmidrule(lr){7-9} \cmidrule(lr){10-12} \cmidrule(lr){13-15}
         & & $k$=0 & $k$=1 & $k$=0 & $k$=1 & Base & +RAG & +PAG & Base & +RAG & +PAG & Base & +RAG & +PAG \\
        \midrule[0.75pt]
        {\tiny\textsc{LaMP-1: Personal.}} & Acc $\uparrow$ & 0.608 & 0.616 & 0.640 & \underline{0.656} & 0.584 & 0.608 & 0.648 & 0.600 & 0.608 & \textbf{0.680} & 0.560 & 0.584 & 0.664 \\
        {\tiny\textsc{Citation Identif.}} & F1 $\uparrow$ & 0.605 & 0.615 & 0.634 & \underline{0.653} & 0.579 & 0.608 & 0.644 & 0.598 & 0.608 & \textbf{0.675} & 0.553 & 0.567 & 0.658 \\
        \midrule
        {\tiny\textsc{LaMP-2: Personal.}} & Acc $\uparrow$ & 0.348 & 0.428 & 0.470 & 0.490 & 0.363 & 0.442 & 0.495 & 0.523 & \underline{0.537} & \textbf{0.551} & 0.463 & 0.467 & 0.507 \\
        {\tiny \textsc{Movie Tagging}} & F1 $\uparrow$ & 0.286 & 0.359 & 0.361 & 0.402 & 0.296 & 0.363 & 0.408 & 0.388 & \underline{0.424} & \textbf{0.442} & 0.320 & 0.349 & 0.385 \\
        \midrule
        {\tiny\textsc{LaMP-3: Personal.}} & MAE $\downarrow$ & 0.291 & \underline{0.252} & 0.274 & 0.270 & 0.503 & 0.408 & 0.408 & 0.272 & \textbf{0.244} & 0.256 & 0.262 & 0.272 & 0.266 \\
        {\tiny\textsc{Product Rating}} & RMSE $\downarrow$ & 0.575 & 0.528 & 0.547 & 0.544 & 0.727 & 0.648 & 0.648 & 0.543 & \textbf{0.510} & \underline{0.522} & 0.558 & 0.580 & 0.561 \\
        \midrule
        {\tiny\textsc{LaMP-4: Personal.}} & R-1 $\uparrow$ & 0.186 & 0.202 & 0.194 & 0.207 & 0.179 & 0.192 & 0.195 & 0.198 & \underline{0.208} & \textbf{0.210} & 0.193 & 0.205 & 0.209 \\
        {\tiny\textsc{News Headline Gen.}} & R-L $\uparrow$ & 0.168 & 0.183 & 0.176 & 0.188 & 0.162 & 0.174 & 0.177 & 0.179 & \underline{0.189} & \textbf{0.191} & 0.173 & 0.185 & 0.190 \\
        \midrule
        {\tiny\textsc{LaMP-5: Personal.}} & R-1 $\uparrow$ & 0.489 & 0.508 & 0.495 & \underline{0.511} & 0.494 & 0.500 & 0.509 & 0.492 & 0.508 & \textbf{0.512} & 0.490 & 0.509 & 0.512 \\
        {\tiny\textsc{Scholarly Title Gen.}} & R-L $\uparrow$ & 0.440 & 0.453 & 0.442 & \underline{0.458} & 0.448 & 0.453 & \textbf{0.460} & 0.443 & 0.454 & \underline{0.458} & 0.439 & 0.457 & 0.459 \\
        \midrule
        {\tiny\textsc{LaMP-7: Personal.}} & R-1 $\uparrow$ & 0.527 & 0.553 & 0.536 & \underline{0.559} & 0.507 & 0.557 & 0.554 & 0.520 & \textbf{0.563} & 0.554 & 0.529 & 0.559 & 0.561 \\
        {\tiny\textsc{Tweet Paraphrasing}} & R-L $\uparrow$ & 0.480 & 0.511 & 0.485 & \underline{0.516} & 0.460 & 0.511 & 0.511 & 0.469 & \textbf{0.521} & 0.512 & 0.480 & 0.515 & 0.519 \\
        \bottomrule[1pt]
    \end{tabular}
    \vspace{-0.05in}
    \caption{\textbf{Main results on LaMP.} R-1 and R-L denote ROUGE-1 and ROUGE-L, respectively. $k$ is the number of retrieved items. For both Per-Pcs and {\sname}, $k$=1 is used for RAG. NP (Non-Personalized) refers to the performance of the task-adapted model. OPPU, training personalized PEFTs directly on user data, can be regarded as an upper bound in case of sufficient data. The highest score is shown in \textbf{bold}, and the second-highest is \underline{underlined}.}
    \label{table:main}
    \vspace{-0.15in}
\end{table*}

\paragraph{Evaluation.}
For utility evaluation, we follow the standard protocol for the benchmark~\citep{salemi2024lamp}, using accuracy and F1 score for text classification tasks (LaMP-1 and LaMP-2), mean absolute error (MAE) and root mean squared error (RMSE) for ordinal multi-class classification tasks (LaMP-3), and ROUGE-1 and ROUGE-L for generation tasks (LaMP-4, LaMP-5, and LaMP-7).
Privacy risks are evaluated via MIA on the personalized modules produced by {\sname} and Per-Pcs, using AUC as the metric, with higher values indicating greater risk of membership inference.

\subsection{Main results}
\label{ss:exp-main}

\paragraph{Overall performance.}
Table~\ref{table:main} summarizes the results across six LaMP tasks, where {\sname} optimizes solely for alignment with target user preferences, i.e., $\alpha$ set to 0.
In classification tasks, {\sname} achieves relative improvements in base performance over Per-Pcs of 2.7\% in accuracy and 3.3\% in F1 score on LaMP-1, and 44.1\% and 31.1\% on LaMP-2, respectively.
For LaMP-3, despite extensive tuning, we observe significantly worse performance for Per-Pcs than the NP baseline, suggesting sensitivity to hyperparameters or limitations of its autoregressive merging.
In contrast, {\sname} improves MAE and RMSE by 45.9\% and 25.3\%, respectively, surpassing also the originally reported Per-Pcs results~\citep{tan2024personalized}.
In generation tasks, {\sname} achieves relative improvements of 10.5\% in ROUGE-1 and 11.1\% in ROUGE-L on LaMP-4, and 2.6\% and 2.0\% on LaMP-7.

{\sname} performs competitively with OPPU~\citep{tan2024democratizing}, a simple approach that directly trains on target user data, often surpassing its performance.
Note that, due to limited reproducibility and missing details, we include in Table~\ref{table:main} the results as originally reported.
On LaMP-3, LaMP-4, and LaMP-5, {\sname} slightly outperforms OPPU in terms of base performance, while on LaMP-7, it marginally lags behind.
More substantial improvements are observed on LaMP-1 and LaMP-2, with relative gains of 7.1\% in accuracy and 8.1\% in F1 score on LaMP-1, and 13.8\% in accuracy and 34.4\% in F1 score on LaMP-2.
These results suggest that strong personalization can be achieved without directly training on target user data, but instead by merging shared LoRA modules.

We observe that PAG, particularly when combined with RAG, performs well on several tasks, including LaMP-1 and LaMP-4. %
This likely arises from the users' consistent behavior patterns, which can be captured succinctly in text.
However, for users with more complex or evolving behaviors, text-based encoding may be less effective.
In such cases, combining prompt- and training-based methods could yield better results, though conflicts between parametric and non-parametric knowledge are often observed in our experiments.

\begin{table*}[t]
    \centering
    \centering
    \footnotesize  %
    \begin{tabular}{lccccccccc}
        \toprule[1pt]
        & \multicolumn{3}{c}{\textbf{LaMP-2}} & \multicolumn{3}{c}{\textbf{LaMP-4}} & \multicolumn{3}{c}{\textbf{LaMP-7}} \\
        \cmidrule(lr){2-4} \cmidrule(lr){5-7} \cmidrule(lr){8-10}
        & Acc $\uparrow$ & F1 $\uparrow$ & AUC $\downarrow$ & R-1 $\uparrow$ & R-L $\uparrow$ & AUC $\downarrow$ & R-1 $\uparrow$ & R-L $\uparrow$ & AUC $\downarrow$ \\
        \midrule[0.75pt]
        Per-Pcs & 0.369 & 0.310 & 0.523 & 0.182 & 0.165 & 0.514 & 0.518 & 0.467 & 0.502 \\
        \midrule[0.75pt]
        $\alpha$=0.0 & 0.523 & 0.388 & 0.539 & \textbf{0.198} & \textbf{0.179} & 0.462 & 0.520 & 0.469 & 0.512 \\
        $\alpha$=0.2 & \textbf{0.547} & \textbf{0.412} & 0.531 & 0.190 & {0.173} & 0.441 & \textbf{0.527} & {0.472} & 0.510 \\
        $\alpha$=0.6 & {0.527} & {0.388} & 0.529 & 0.186 & 0.168 & {0.412} & {0.524} & \textbf{0.476} & 0.501 \\
        \rowcolor{RowHighlight} $\alpha$=1.0 & 0.524 & 0.385 & {0.522} & 0.185 & 0.168 & \textbf{0.408} & 0.522 &{0.472} & {0.499} \\
        \rowcolor{RowHighlight} $\alpha$=2.0 & 0.503 & 0.368 & \textbf{0.507} & 0.182 & 0.166 & \textbf{0.408} & 0.519 & 0.469 & \textbf{0.498} \\
        \bottomrule[1pt]
    \end{tabular}
    \vspace{-0.05in}
    \captionof{table}{\textbf{Privacy-utility trade-off with varying \bm{$\alpha$}.} Overall, as $\alpha$ increases, task utility metrics decrease, and MIA effectiveness also drops. $\alpha$ values of 1.0 and 2.0 achieve a strictly better privacy-utility trade-off than Per-Pcs.}
    \label{table:main-table-p}
    \vspace{-0.05in}
\end{table*}

\begin{table*}[ht]
    \centering
    \footnotesize  %
    \setlength{\tabcolsep}{5pt}
    \begin{tabular}[b]{p{0.8\linewidth}}
        \toprule[1pt]
        \textbf{Prompt:} Generate a headline for the following article: But it got us thinking about the concept of a ``healthier'' option -- is it always, in fact, better for you? In some instances \\
        \midrule[0.25pt]
        \textbf{$\bm{s_1}$:} \colorbox{yellow!30}{Healthier Food Swaps That Aren't Really Healthier (PHOTOS)} \\
        \midrule[0.25pt]
        \textbf{$\bm{\alpha}$=0.0:} \colorbox{yellow!30}{Healthier Foods That Aren't} Actually \colorbox{yellow!30}{Healthier (PHOTOS)} \\ %
        \midrule[0.25pt]
        \textbf{$\bm{\alpha}$=5.0:} 10 `Healthy' \colorbox{yellow!30}{Foods That} Are Actually Bad For You \\ %
        \midrule[0.75pt]
        \textbf{Prompt:} Generate a headline for the following article: ``If you leave the house without sunscreen, you might as well be naked.'' -- Meghan McCain ``We talk about skin health, ...'' \\
        \midrule[0.25pt]
        \textbf{$\bm{s_1}$:} \colorbox{yellow!30}{Meghan McCain, Kate Upton And More Celebrities Reveal Their Beauty Secrets (PHOTOS)} \\
        \midrule[0.25pt]
        \textbf{$\bm{\alpha}$=0.0:} \colorbox{yellow!30}{Meghan McCain}'s Skin Care Routine: `I'm Not Afraid To Try New Things' (VIDEO) \\ %
        \midrule[0.25pt]
        \textbf{$\bm{\alpha}$=5.0:} The Best Beauty Advice From The Women Of The View \\ %
        \bottomrule[1pt]
    \end{tabular}
    \vspace{-0.05in}
    \captionof{table}{\textbf{Response comparison with varying $\bm{\alpha}$.} On LaMP-4, at $\alpha = 0.0$, responses are more aligned with the top sharing user $s_1$ than at $\alpha = 5.0$, where greater lexical and phrasal differences appear.}
    \label{table:privacy-demo-nh-s}
    \vspace{-0.10in}
\end{table*}

\paragraph{Privacy-utility trade-off analysis.}

We evaluate {\sname} in terms of the privacy-utility trade-off it achieves compared to Per-Pcs under varying privacy levels across three benchmark tasks: LaMP-2, LaMP-4, and LaMP-7.
Utility is measured using two task-specific metrics, and privacy risk is assessed by MIA performance, measured in terms of the AUC score, where lower values indicate stronger privacy.
As shown in Table~\ref{table:main-table-p}, {\sname} consistently outperforms Per-Pcs in utility, achieving comparable or better results at every $\alpha$ value considered.
As $\alpha$ increases, utility generally declines, with a similar reduction in MIA risk.
Notably, at $\alpha$ values of 1.0 and 2.0, {\sname} achieves a strictly better privacy-utility trade-off compared to Per-Pcs, with a 20.6\% reduction in MIA risk while maintaining comparable utility on LaMP-4 at $\alpha$ of 2.0.
These results demonstrate that our method enables a controllable privacy-utility trade-off, outperforming the baseline for appropriate values of $\alpha$.
Note that while AUC values near 0.5 generally suggest limited privacy leakage, the key result is that {\sname} consistently lowers MIA risks compared to Per-Pcs while achieving superior utility.
Extending the benchmark to enable more comprehensive and realistic MIA evaluations remains an important direction for future work.

Table~\ref{table:privacy-demo-nh-s} presents additional qualitative examples comparing responses generated with LoRA modules merged at different $\alpha$ values to those of the most similar sharing user.
At $\alpha$ of 0.0, the responses closely resemble those of the top sharing user, with similar sentence structure and word choices.
In contrast, at $\alpha$ of 5.0, the responses become more distinct.
These examples illustrate how varying $\alpha$ values control the degree of response similarity with the sharing user.

\subsection{Ablations and analysis}
\label{ss:ablations}

\paragraph{Use of RAG and PAG.}

\begin{figure}[t]
    \centering
    \begin{subfigure}[t]{0.49\linewidth}
        \centering
        \includegraphics[width=\linewidth]{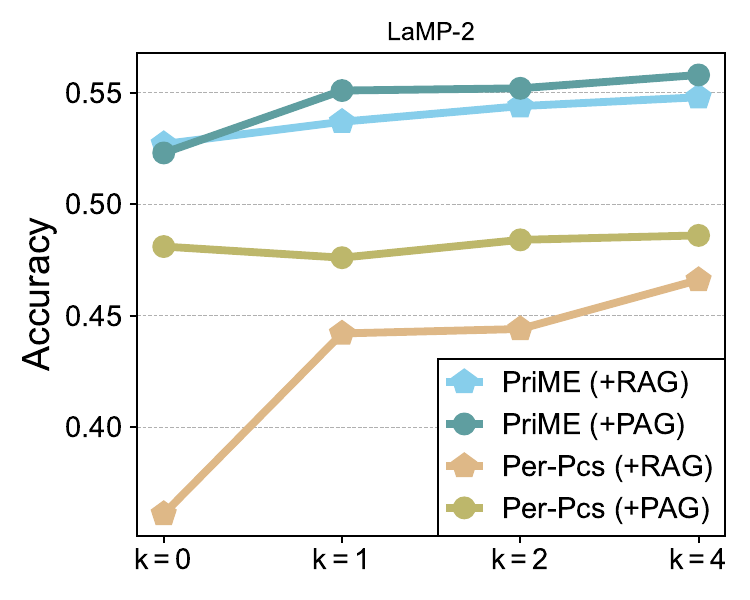}
    \end{subfigure}
    \hfill
    \begin{subfigure}[t]{0.49\linewidth}
        \centering
        \includegraphics[width=\linewidth]{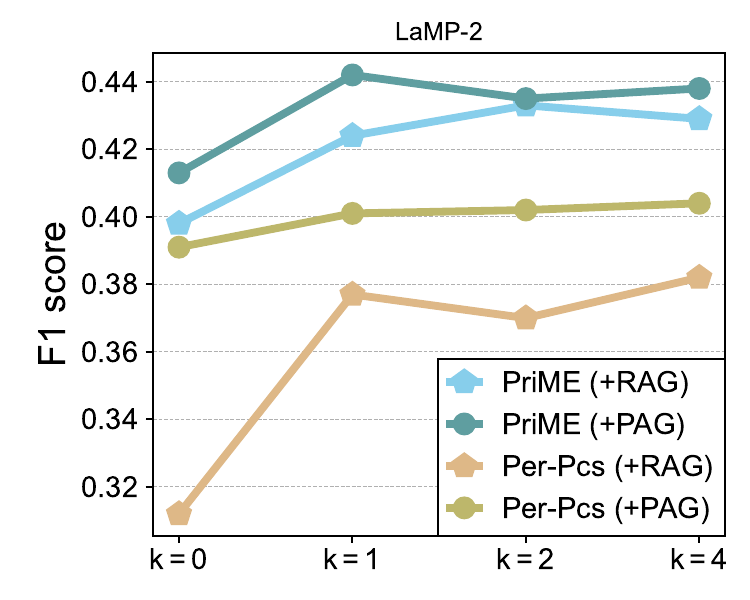}
    \end{subfigure}
    \\
    \centering
    \begin{subfigure}[t]{0.49\linewidth}
        \centering
        \includegraphics[width=\linewidth]{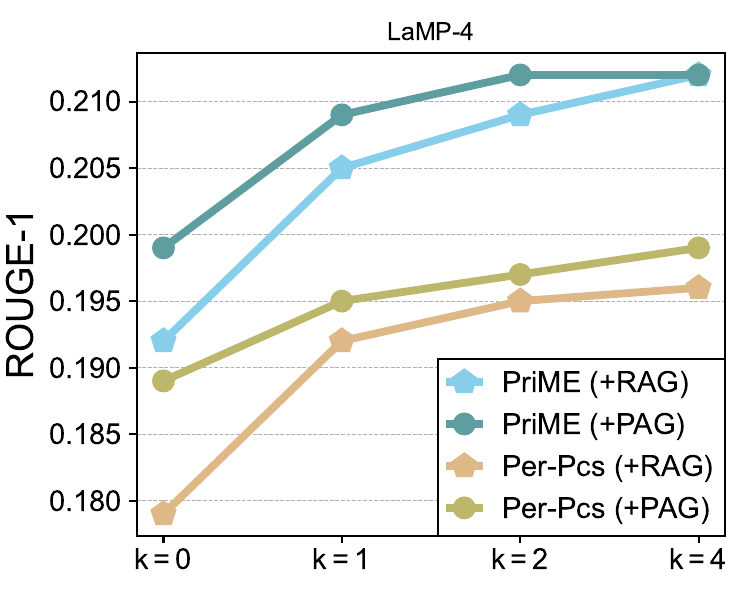}
    \end{subfigure}
    \hfill
    \begin{subfigure}[t]{0.49\linewidth}
        \centering
        \includegraphics[width=\linewidth]{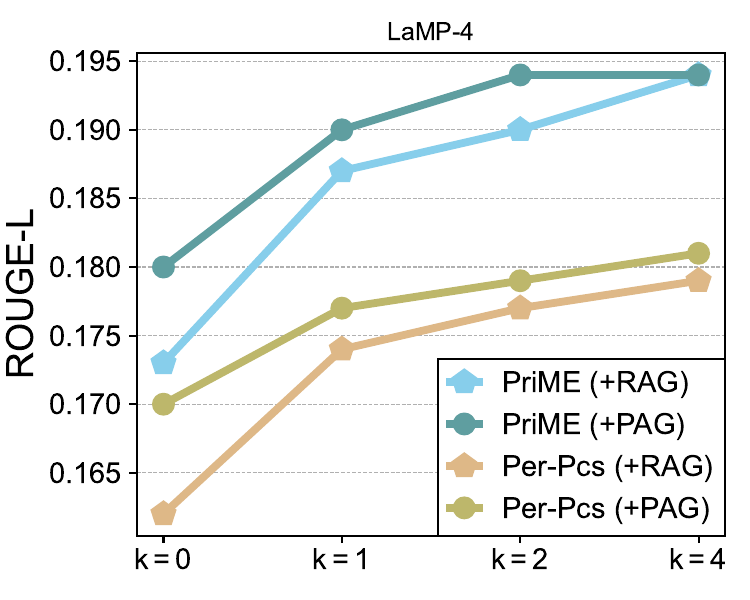}
    \end{subfigure}
    \caption{\textbf{Performance comparison with varying $\bm{k}$ in RAG, with and without PAG}. Overall performance improves with increasing $k$ and profile augmentation for both {\sname} and Per-Pcs. Per-Pcs benefits more from RAG and PAG, showing greater reliance on the augmentation methods to achieve competitive results.}
    \vspace{-0.05in}
    \label{fig:rag-pag}
\end{figure}

We evaluate the impact of varying the number of retrieved items ($k$) in RAG on the effectiveness of {\sname} and Per-Pcs, and assess how PAG further affects performance.
As shown in Figure~\ref{fig:rag-pag}, performance improves with $k$ for both methods, though the gains plateau beyond $k$ at 2, with additional benefits from profile augmentation.
Per-Pcs shows greater dependence on RAG and PAG for competitive performance, while {\sname} sees more moderate improvements, especially on LaMP-2.
Notably, {\sname} without RAG outperforms Per-Pcs with RAG at $k$ of 4 on LaMP-2, indicating that a well-personalized module can reduce the need for prompt augmentation.

\paragraph{LoRA count and merging granularity.}

\begin{table}[t]
    \centering
    \scriptsize  %
    \begin{tabular}{lcccccc}
        \toprule[1pt]
        & \multicolumn{2}{c}{\textbf{LaMP-2}} & \multicolumn{2}{c}{\textbf{LaMP-4}} & \multicolumn{2}{c}{\textbf{LaMP-7}}\\
        \cmidrule(lr){2-3} \cmidrule(lr){4-5} \cmidrule(lr){6-7}
        & Acc $\uparrow$ & F1 $\uparrow$ & R-1 $\uparrow$ & R-L $\uparrow$ & R-1 $\uparrow$ & R-L $\uparrow$ \\
        \midrule[0.75pt]
        Per-Pcs & 0.363 & 0.296 & 0.179 & 0.162 & 0.507 & 0.460 \\
        \midrule[0.75pt]
        $n=1$ & 0.501 & 0.372 & 0.193 & 0.175 & \textbf{0.523} & \textbf{0.470} \\
        $n=3$ & \textbf{0.523} & \textbf{0.388} & \textbf{0.198} & \textbf{0.179} & 0.520 & 0.469 \\
        $n=5$ & \textbf{0.523} & \textbf{0.388} & 0.195 & 0.177 & 0.517 & 0.465 \\
        \bottomrule[1pt]
    \end{tabular}
    \vspace{-0.05in}
    \caption{\textbf{Performance with varying number of LoRAs.} Performance often improves with more LoRAs, with a single LoRA in {\sname} often outperforming Per-Pcs.}
    \label{table:lora-count}
    \vspace{-0.05in}
\end{table}

We evaluate {\sname} on LaMP-2, LaMP-4, and LaMP-7 with different numbers of LoRAs used for merging.
As shown in Table~\ref{table:lora-count}, performance often improves with more LoRAs.
Notably, using only a single LoRA, which produces a target user module simply by scaling the selected LoRA, still outperforms Per-Pcs, which performs merging at the layer level.
This shows that effective personalization is possible with coarser LoRA-level merging, provided interpolation weights are appropriately optimized.

\begin{figure}[t]
    \centering
    \begin{subfigure}[t]{0.49\linewidth}
        \centering
        \includegraphics[width=\linewidth]{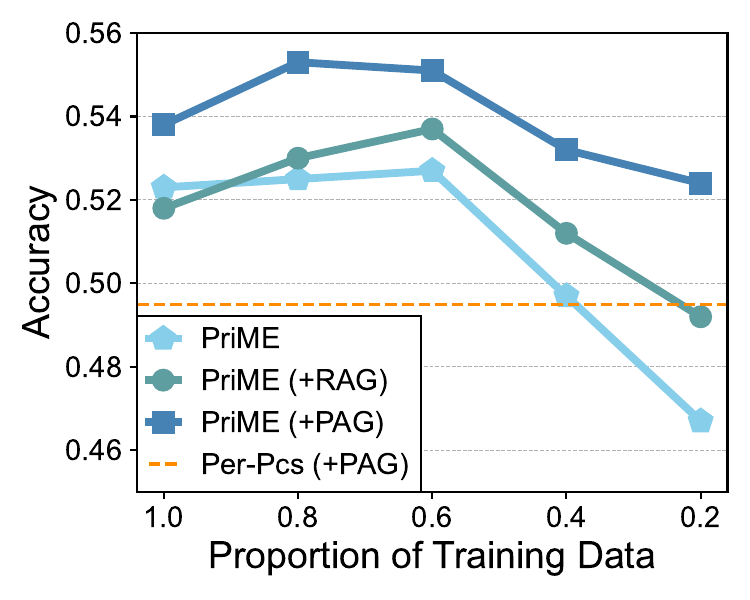}
    \end{subfigure}
    \hfill
    \begin{subfigure}[t]{0.49\linewidth}
        \centering
        \includegraphics[width=\linewidth]{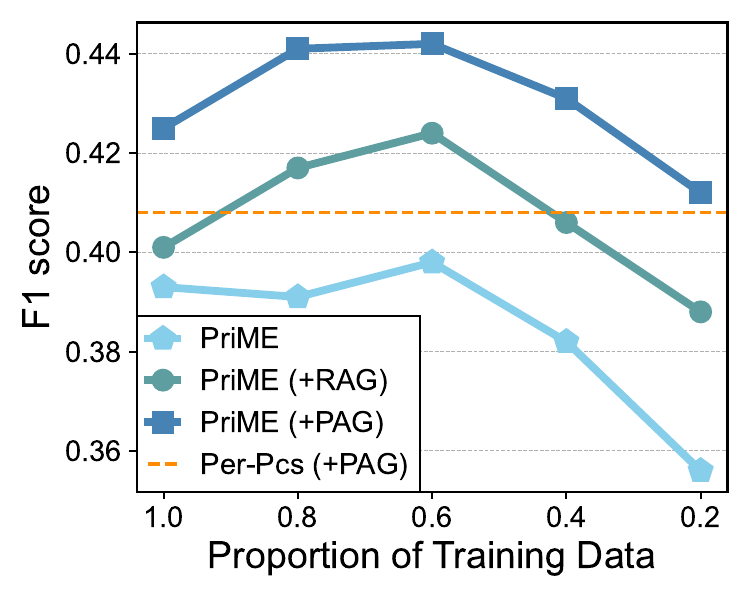}
    \end{subfigure}
    \caption{\textbf{Performance with varying proportions of training data used for optimization.} The dotted line represents the best performance achieved by Per-Pcs using the full training data.}
    \vspace{-0.05in}
    \label{fig:train-data-sample}
\end{figure}

\paragraph{Data sampling for efficiency.}
While gradient-free methods enable direct optimization of arbitrary objectives, computing metrics such as ROUGE requires full language model decoding, resulting in substantial computational cost at scale.
To improve efficiency, we evaluate {\sname} with varying fractions of the data used in merging.
As illustrated in Figure~\ref{fig:train-data-sample}, on LaMP-2, training with only 60\% of the data achieves performance comparable to that of using the full dataset.
Moreover, with only 20\% of the data, we surpass the best-performing variant of Per-Pcs, which is with PAG.
These results indicate that significant reductions in training data are possible for {\sname} without compromising effectiveness, enhancing the scalability of the method.

\paragraph{Optimizer comparison.}

\begin{figure}[t]
    \centering
    \begin{subfigure}[t]{0.49\linewidth}
        \centering
        \includegraphics[width=\linewidth]{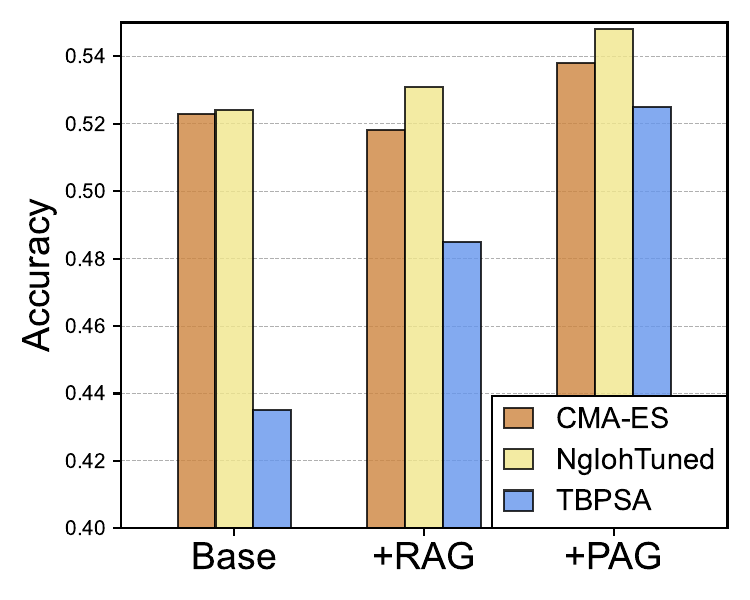}
    \end{subfigure}
    \hfill
    \begin{subfigure}[t]{0.49\linewidth}
        \centering
        \includegraphics[width=\linewidth]{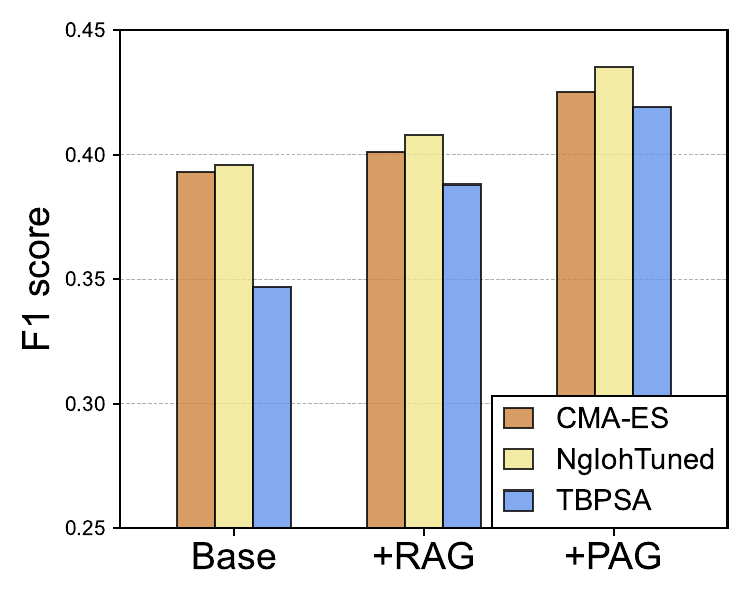}
    \end{subfigure}
    \caption{\textbf{Comparison of different gradient-free optimization algorithms in {\sname}.} General-purpose optimizers like CMA-ES provide competitive results on the benchmark, while specialized optimizers such as TBPSA may be considered in specific scenarios, such as when strong noise is present in evaluating $f$.}
    \vspace{-0.05in}
    \label{fig:optimizers}
\end{figure}

Although {\sname} is agnostic to the choice of gradient-free optimzer, we evaluate several popular optimizers and their impact on performance.
Specifically, we consider three optimizers: vanilla CMA-ES, a widely used evolutionary strategy; NgIohTuned~\citep{bennet2021nevergrad}, a meta-optimizer that dynamically selects among strategies (including CMA-ES variants) depending on the problem setting; and TBPSA~\citep{hellwig2016evolution}, which adapts population size to better handle potentially strong noise in the fitness function.
As shown in Figure~\ref{fig:optimizers}, general-purpose optimizers such as CMA-ES and NgIohTuned perform well on LaMP-2, while TBPSA yields comparatively worse results.
Although language model decoding introduces some stochasticity, the evaluation function $f$ is deterministic in our setting.
Since evolutionary algorithms are generally robust to moderate noise in the objective function~\citep{arnold2012noisy}, highly specialized optimizers like TBPSA may not be the most suitable in this context.
Nevertheless, using gradient-free methods allows more flexibility across different conditions, improving broad applicability.

\begin{table*}[ht]
    \centering
    \scriptsize  %
    \setlength{\tabcolsep}{5pt}
    \begin{subtable}[t]{0.49\textwidth}
    \begin{tabular}{llcccccc}
        \toprule[1pt]
        \multirow{2.5}{*}{\textbf{Task}} & \multirow{2.5}{*}{\textbf{Metric}} & \multicolumn{3}{c}{\textbf{\textsc{Per-Pcs}}} & \multicolumn{3}{c}{\textbf{\textsc{\sname} (Ours)}} \\ 
        \cmidrule(lr){3-5} \cmidrule(lr){6-8}
         & & Base & +RAG & +PAG & Base & +RAG & +PAG \\
        \midrule[0.75pt]
        {\tiny LaMP-2} & Acc $\uparrow$ & 0.381 & 0.436 & 0.473 & 0.514 & \underline{0.534} & \textbf{0.545} \\
         & F1 $\uparrow$ & 0.338 & 0.374 & 0.405 & 0.408 & \underline{0.425} & \textbf{0.434} \\
        \midrule
        {\tiny LaMP-4} & R-1 $\uparrow$ & 0.196 & 0.205 & 0.209 & 0.203 & \underline{0.213} & \textbf{0.218} \\
         & R-L $\uparrow$ & 0.176 & 0.185 & 0.188 & 0.184 & \underline{0.194} & \textbf{0.198} \\
        \bottomrule[1pt]
    \end{tabular}
    \caption{Llama3.1-8B}
    \end{subtable}
    \hfill
    \begin{subtable}[t]{0.49\textwidth}
    \begin{tabular}{llcccccccccc}
        \toprule[1pt]
        \multirow{2.5}{*}{\textbf{Task}} & \multirow{2.5}{*}{\textbf{Metric}} & \multicolumn{3}{c}{\textbf{\textsc{Per-Pcs}}} & \multicolumn{3}{c}{\textbf{\textsc{\sname} (Ours)}} \\ 
        \cmidrule(lr){3-5} \cmidrule(lr){6-8}
         & & Base & +RAG & +PAG & Base & +RAG & +PAG \\
        \midrule[0.75pt]
        {\tiny LaMP-2} & Acc $\uparrow$ & 0.327 & 0.402 & 0.462 & 0.496 & \underline{0.515} & \textbf{0.527} \\
         & F1 $\uparrow$ & 0.278 & 0.349 & 0.391 & 0.366 & \underline{0.411} & \textbf{0.413} \\
        \midrule
        {\tiny LaMP-4} & R-1 $\uparrow$ & 0.180 & 0.191 & 0.190 & 0.184 & \underline{0.197} & \textbf{0.202} \\
         & R-L $\uparrow$ & 0.162 & 0.173 & 0.172 & 0.165 & \underline{0.180} & \textbf{0.183} \\
        \bottomrule[1pt]
    \end{tabular}
    \caption{Llama3.2-3B}
    \end{subtable}
    \vspace{-0.05in}
    \caption{\textbf{LaMP results for Llama 3 models.} For both the 8B and 3B models, {\sname} outperforms Per-Pcs on LaMP-2 (multi-class classification) and LaMP-4 (generation) tasks.}
    \label{table:app-l3}
\end{table*}

\begin{table*}[t]
    \centering
    \centering
    \scriptsize  %
    \begin{subtable}[t]{0.49\textwidth}
    \begin{tabular}{lcccccc}
        \toprule[1pt]
        & \multicolumn{3}{c}{\textbf{LaMP-2}} & \multicolumn{3}{c}{\textbf{LaMP-4}} \\
        \cmidrule(lr){2-4} \cmidrule(lr){5-7}
        & Acc $\uparrow$ & F1 $\uparrow$ & AUC $\downarrow$ & R-1 $\uparrow$ & R-L $\uparrow$ & AUC $\downarrow$ \\
        \midrule[0.75pt]
        Per-Pcs & 0.381 & 0.338 & 0.526 & 0.196 & 0.176 & 0.526 \\
        \midrule[0.75pt]
        $\alpha$=0.0 & 0.514 & 0.408 & 0.523 & \textbf{0.203} & \textbf{0.184} & 0.467 \\
        $\alpha$=0.2 & 0.520 & \textbf{0.414} & 0.517 & 0.201 & 0.182 & 0.452 \\
        $\alpha$=0.6 & 0.510 & 0.401 & 0.514 & 0.195 & 0.175 & 0.421 \\
        $\alpha$=1.0 & \textbf{0.526} & \textbf{0.414} & 0.513 & 0.192 & 0.173 & 0.416 \\
        $\alpha$=2.0 & 0.513 & 0.396 & \textbf{0.503} & 0.193 & 0.174 & \textbf{0.413} \\
        \bottomrule[1pt]
    \end{tabular}
    \caption{Llama3.1-8B}
    \end{subtable}
    \hfill
    \begin{subtable}[t]{0.49\textwidth}
    \begin{tabular}{lcccccc}
        \toprule[1pt]
        & \multicolumn{3}{c}{\textbf{LaMP-2}} & \multicolumn{3}{c}{\textbf{LaMP-4}} \\
        \cmidrule(lr){2-4} \cmidrule(lr){5-7}
        & Acc $\uparrow$ & F1 $\uparrow$ & AUC $\downarrow$ & R-1 $\uparrow$ & R-L $\uparrow$ & AUC $\downarrow$ \\
        \midrule[0.75pt]
        Per-Pcs & 0.327 & 0.278 & 0.530 & 0.180 & 0.162 & 0.497 \\
        \midrule[0.75pt]
        $\alpha$=0.0 & \textbf{0.496} & \textbf{0.366} & 0.524 & \textbf{0.184} & \textbf{0.165} & 0.451 \\
        $\alpha$=0.2 & 0.493 & 0.360 & 0.517 & 0.177 & 0.159 & 0.433 \\
        $\alpha$=0.6 & 0.484 & 0.359 & 0.518 & 0.171 & 0.153 & 0.405 \\
        $\alpha$=1.0 & 0.479 & 0.349 & 0.503 & 0.168 & 0.152 & 0.402 \\
        $\alpha$=2.0 & 0.458 & 0.330 & \textbf{0.499} & 0.168 & 0.151 & \textbf{0.400} \\
        \bottomrule[1pt]
    \end{tabular}
    \caption{Llama3.2-3B}
    \end{subtable}
    \vspace{-0.05in}
    \caption{\textbf{Privacy-utility trade-offs for Llama 3 models.} {\sname} achieves strictly better trade-offs for multiple values of $\alpha$ on LaMP-2 and LaMP-4 for both the 8B and 3B models.}
    \label{table:app-l3-p}
    \vspace{-0.05in}
\end{table*}

\paragraph{Alternative base models.}

To assess how {\sname} performs across model architectures and sizes, we conduct additional experiments with the Llama 3 family of models~\citep{grattafiori2024llama} on LaMP-2 and LaMP-4, specifically using the Llama3.1-8B and Llama3.2-3B models.
Table~\ref{table:app-l3} summarizes the results for the two models with {\sname} optimizing solely for alignment with target user preferences, i.e., $\alpha$ set to 0.
Similar to the results for Llama-2-7b, {\sname} demonstrates superior performance on both the multi-class classification (LaMP-2) and generation (LaMP-4) tasks compared to Per-Pcs.
On LaMP-2, {\sname} achieves relative improvements over Per-Pcs in base performance with gains of 35.0\% in accuracy and 20.7\% in F1 score for the 8B model, and 51.7\% in accuracy and 31.7\% in F1 score for the 3B model.
On LaMP-4, {\sname} achieves improvements of 3.3\% in ROUGE-1 and 4.5\% in ROUGE-L for the 8B model, and 2.2\% in ROUGE-1 and 1.9\% in ROUGE-L for the 3B model.
Notably, Llama3.2-3B with {\sname} outperforms Llama3.1-8B with Per-Pcs on LaMP-2.

Regarding privacy-utility trade-offs, Table~\ref{table:app-l3-p} shows the trade-offs achieved by Per-Pcs and {\sname} for varying values of $\alpha$.
On LaMP-2, {\sname} consistently achieves strictly better trade-offs across all $\alpha$ values, with AUC improving as $\alpha$ increases.
On LaMP-4, {\sname} results in better trade-offs for $\alpha = 0.0$ and $\alpha = 0.2$ for the 8B model, and for $\alpha = 0.0$ for the 3B model.

\section{Related Work}
\label{s:related}

\paragraph{LLM personalization.}
Personalization aims to adapt general-purpose models to individual user preferences.
Prompt-based methods incorporate user preferences into input prompts, enabling personalization without additional training~\citep{mysore2024pearl,richardson2023integrating,kim2025few}.
Approaches range from in-context learning with few-shot user examples~\citep{christakopoulou2023large} to retrieval-augmented generation (RAG)~\citep{salemi2024lamp}, which integrates relevant user records into prompts, and profile-augmented generation (PAG)~\citep{richardson2023integrating}, which leverages abstract user profile summaries to enhance query relevance.
While simple to implement, these methods are constrained by the model's context length and incur higher inference costs due to enlarged prompts.
Moreover, including potentially sensitive user information as raw text in a prompt can introduce privacy risks.

Training-based methods, on the other hand, directly encode user preferences into model parameters.
OPPU~\citep{tan2024democratizing} is a simple method that fine-tunes a LoRA~\citep{hu2022lora} module per user, but performance may degrade when only limited user data is available for training.
To address this, methods such as group preference optimization~\citep{zhao2024group} and soft prompting~\citep{li2024steerability} leverage data from multiple users or groups.
However, accessing such collective data is rarely available in practice and can raise privacy concerns without explicit safeguards.

Model merging offers alternative methods.
Per-Pcs~\citep{tan2024personalized} merges PEFT modules each trained on different user data to create a personalized module for a target user, enabling more effective personalization when target user data is limited.
However, it introduces additional model complexity, does not optimize for task-specific utility, and fails to explicitly address the privacy risks for sharing users.
In contrast, {\sname} employs evolutionary methods to optimize both capturing user preferences and reducing privacy risks for sharing users in merging-based personalization.

\paragraph{Model merging.}
Model merging is a surprisingly effective approach for developing models by integrating diverse models with complementary capabilities into a unified architecture.
Simple techniques, such as averaging the weights of fine-tuned models, have shown strong performance on tasks such as image classification~\citep{wortsman2022model}.
In language models, methods such as task arithmetic~\citep{ilharco2023editing} construct task vectors that represent model weights encoding task-specific abilities, which are then combined to create a model with the desired capabilities.
Despite its effectiveness, model merging often relies on human intuition and expertise for selecting models and designing merging recipes.
Recently, evolutionary algorithms have been applied to automatically discover effective merging strategies in developing foundation models with diverse capabilities~\citep{akiba2025evolutionary}.
In this work, we build on this approach by applying evolutionary methods to personalize language models, with an explicit optimization objective for reducing privacy risks associated with shared user data.

\section{Conclusion}
\label{s:conclusion}

In this work, we introduce {\sname}, a novel approach to LLM personalization that leverages evolutionary algorithms to directly optimize task utility while mitigating privacy risks for users sharing their data.
Experiments on LaMP demonstrate that {\sname} outperforms competing methods, achieving substantial improvements in capturing user preferences and enhanced robustness to membership inference attacks.
These findings highlight the potential of evolutionary model merging as a promising framework for privacy-aware LLM personalization.

\paragraph{Future work.}
Further exploration of advanced model merging techniques---such as random weight dropping~\citep{yu2024language} and the removal of conflicting weights~\citep{yadav2024ties}---and their impact on the resulting personalized modules and privacy risks associated with sharing users would be a valuable direction for future work.
Additionally, addressing conflicts between parametric and non-parametric knowledge when combining prompt-based and training-based methods, or between two sources of non-parametric knowledge (e.g., RAG and PAG), as observed in our empirical evaluation, could provide deeper insights into further enhancing LLM personalization.

\section*{Limitations}

While our proposed {\sname} demonstrates promising results in LLM personalization, further investigation is needed to assess how well it scales with larger models and datasets.
Our experimental results suggest that {\sname} remains effective even with substantially reduced training data.
However, developing more sophisticated strategies for selecting the most useful subset of data for merging could further improve its scalability.
Moreover, while our evaluation shows that the approach empirically mitigates MIA risks, it does not offer formal privacy guarantees.
Future work should explore integrating rigorous mechanisms such as differential privacy~\citep{dwork2006differential}.

\section*{Ethics Statement}
Personalization in LLMs aims to adapt general-purpose models to capture the preferences of individual users or groups. 
This often involves access to personal data, which may contain sensitive information, making privacy protection crucial.
While our evaluation shows that {\sname} achieves effective personalization and reduces the risk of membership inference attacks, emerging threats---such as LLMs inferring personal attributes from seemingly innocuous text~\citep{staab2024beyond,kim2025self}---remain a concern. 
Future research on personalization should carefully consider such emerging privacy risks.

We used an AI assistant (ChatGPT) to refine the writing during the preparation of this work.
All models and datasets are publicly available and used in accordance with their intended purpose; details are provided in Appendix~\ref{s:appendix:exp-details}.

\section*{Acknowledgments}
This research was supported in part by Institute for Information \& communications Technology Planning \& Evaluation (IITP) grant funded by the Korea government (MSIT) (No. RS-2019-II190075, Artificial Intelligence Graduate School Support Program (KAIST); No. RS-2020-II201361, Artificial Intelligence Graduate School Program (Yonsei University); No. RS-2021-II212068, Artificial Intelligence Innovation Hub; No. RS-2025-02215344, Development of AI Technology with Robust and Flexible Resilience Against Risk Factors) and the Mobile eXperience (MX) Business, Samsung Electronics Co., Ltd.

\bibliography{custom}

\clearpage
\appendix
\section{Benchmark and Task Details}
\label{s:appendix:benchmark-details}

In this section, we provide additional details on the individual tasks in the benchmark.

\paragraph{LaMP-1: Personalized citation identification.}
This is a binary classification task in which a language model identifies which of two candidate papers a user would cite, given the user's history of previous publications.
User's profile contains a set of titles, abstracts, and citations of previous publications.
During task adaptation, the model is trained to generate a citation based on the title of a paper.

\paragraph{LaMP-2: Personalized movie tagging.}
In this multi-class classification task, a language model needs to classify a given movie description into one of 15 possible categories, e.g., `comedy' or `action'.
Given a user's historical tagging assignments, the model is adapted to generate appropriate tags for movie descriptions.

\paragraph{LaMP-3: Personalized product rating.}
In this ordinal multi-class classification task, a language model predicts product ratings based on a user's review history.
Given pairs of a review and the corresponding historical rating, the model needs to generate an integer rating from 1 to 5.

\paragraph{LaMP-4: Personalized news headline generation.}
In this task, a language model generates a news headline for an article based on a user's stylistic patterns as an author.
Based on the user's article history and titles, the model is adapted to generate headlines that align with the identified style.

\paragraph{LaMP-5: Personalized scholarly title generation.}
Similar to LaMP-4, in this task, a language model generates research article titles based on a user's profile of historical article-title pairs.
In LaMP-5, only the abstracts of articles are provided.

\paragraph{LaMP-7: Personalized tweet paraphrasing.}
This is another personalized generation task in which a language model needs to paraphrase a tweet in the style of a user based on the user's historical tweets.

\section{Experimental Details}
\label{s:appendix:exp-details}

\begin{table*}[ht]
    \centering
    \footnotesize  %
    \setlength{\tabcolsep}{5pt}
    \begin{tabular}{lcccccccccccccc}
        \toprule[1pt]
        \multirow{2.5}{*}{\textbf{Task}} & \multicolumn{3}{c}{\textbf{Task Adaptation}} & \multicolumn{3}{c}{\textbf{Sharer PEFT}} & \multicolumn{3}{c}{\textbf{Sharer Gate}} & \multicolumn{2}{c}{\textbf{\textsc{Per-Pcs}}} & \multicolumn{3}{c}{\textbf{\textsc{\sname}}} \\ 
        \cmidrule(lr){2-4} \cmidrule(lr){5-7} \cmidrule(lr){8-10} \cmidrule(lr){11-12} \cmidrule(lr){13-15}
         & batch & ep & lr & batch & ep & lr & batch & step & lr & top-$k$ & batch & LoRA & budget & data \% \\
        \midrule[0.75pt]
        \textsc{LaMP-1} & 16 & 3 & 1e-4 & 8 & 1 & 1e-4 & 6 & 100 & 1e-5 & 1 & 16 & 5 & 50 & 0.8 \\
        \midrule
        \textsc{LaMP-2} & 16 & 3 & 1e-4 & 6 & 3 & 2e-4 & 6 & 100 & 2e-5 & 3 & 16 & 3 & 30 & 0.6 \\
        \midrule
        \textsc{LaMP-3} & 8 & 3 & 1e-4 & 4 & 2 & 1e-4 & 4 & 100 & 1e-5 & 1 & 6 & 5 & 50 & 1.0 \\
        \midrule
        \textsc{LaMP-4} & 8 & 3 & 1e-4 & 8 & 3 & 1e-4 & 6 & 50 & 2e-5 & 1 & 16 & 3 & 30 & 0.8 \\
        \midrule
        \textsc{LaMP-5} & 8 & 3 & 1e-4 & 8 & 2 & 1e-4 & 6 & 50 & 2e-5 & 1 & 10 & 3 & 50 & 1.0 \\
        \midrule
        \textsc{LaMP-7} & 16 & 3 & 1e-4 & 8 & 1 & 1e-4 & 6 & 50 & 2e-5 & 2 & 16 & 3 & 30 & 1.0 \\
        \bottomrule[1pt]
    \end{tabular}
    \vspace{-0.05in}
    \caption{\textbf{Hyperparameters.} Hyperparameters for task adaptation, PEFT training, Per-Pcs, and {\sname} on LaMP.}
    \label{table:hparams}
    \vspace{-0.10in}
\end{table*}

\subsection{Setup details}
\label{ss:appendix:setup}

\paragraph{Data splits.}
Following a similar approach to that from prior work~\citep{tan2024personalized}, we split the data by using 25\% of the users for adapting the base model, 100 users for testing, and the rest as candidates for sharing their LoRA modules.

\paragraph{Sharing user selection.}
To select sharing users, we first compute user embeddings with the DeBERTa V3 large model~\citep{he2023debertav3} (see Section~\ref{ss:prime}) and apply $k$-means clustering to group them into up to $50$ clusters.
From each cluster, we select the user with the largest history as the sharing user.
For each selected user, we train a LoRA module using the task-adapted model as the base.

\paragraph{User profile generation.}
User profiles are generated by summarizing a random sample of the user's history using Mistral-7B-Instruct v0.2~\citep{jiang2023mistral}.
These summaries capture user preferences or behavior patterns, e.g., the user's most frequently explored research topics, in text.

\paragraph{Baselines.}
As prior work~\citep{tan2024personalized} does not report task adaptation details, we perform additional hyperparameter tuning to achieve comparable results on LaMP-2, LaMP-4, LaMP-5, and LaMP-7, and improved results on LaMP-1 and LaMP-3.
For Per-Pcs, we follow the reported settings and also conduct additional hyperparameter tuning, reporting the best-performing results.

\paragraph{{\sname}.}
For {\sname}, we optimize the sum of the two utility metrics for each task.
Note that for LaMP-1, the training data consists of textual citations paired with titles and abstracts, differing from the evaluation task, which is a binary classification between two citation options.
Hence, for LaMP-1, we optimize the sum of ROUGE-1 and ROUGE-L scores with the target citations.

\paragraph{MIA.}
We adopt standard MIA setups, assuming an adversary with black-box access to the personalized LLM who queries it with input-response pairs and uses the resulting losses as scores~\citep{yeom2018privacy,duan2024membership}.
The adversary then predicts whether an example was used to train a LoRA module for a sharing user involved in merging.
To compare MIA risks, we evaluate the model’s loss on each sharing user’s training data, treating the data from selected users as members and that from non-selected users as non-members, and compute the AUC to measure the risk of MIA attacks.
We report the average AUC across the 100 test users.

\subsection{Training details}
\label{ss:appendix:train}

\paragraph{Hyperparameters.}
Table~\ref{table:hparams} summarizes the hyperparameters used for task adaptation, LoRA training for sharing users, Per-Pcs, and {\sname}.
For LoRA training, we used the LoRA module hyperparameters reported in \citet{tan2024personalized}, but additionally experimented with training hyperparameters such as learning rate and batch size.
The configurations achieving the best performance on the target metrics were selected.

We note that while we successfully reproduced most of the Per-Pcs results, our LaMP-3 results were noticeably lower despite extensive efforts.
This suggests that Per-Pcs may be more sensitive to hyperparameter choices on certain tasks.
Nevertheless, {\sname} still outperforms the originally reported Per-Pcs results.

\paragraph{Compute resources.}
We conducted our experiments mainly on NVIDIA A6000 GPUs, requiring one GPU to train a single model with LoRA.

\section{Additional Results}
\label{s:appendix:add-results}

This section provides additional experimental results, analysis, and discussion.

\subsection{Sampling training data}
\label{ss:appendix:add-results-sampling}

\begin{figure}[t]
    \centering
    \begin{subfigure}[t]{0.48\textwidth}
        \centering
        \includegraphics[width=0.49\linewidth]{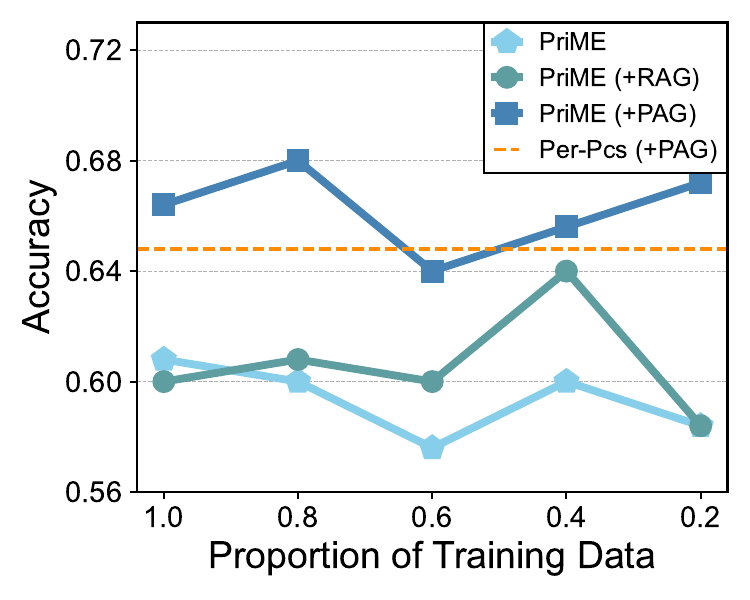}
        \includegraphics[width=0.49\linewidth]{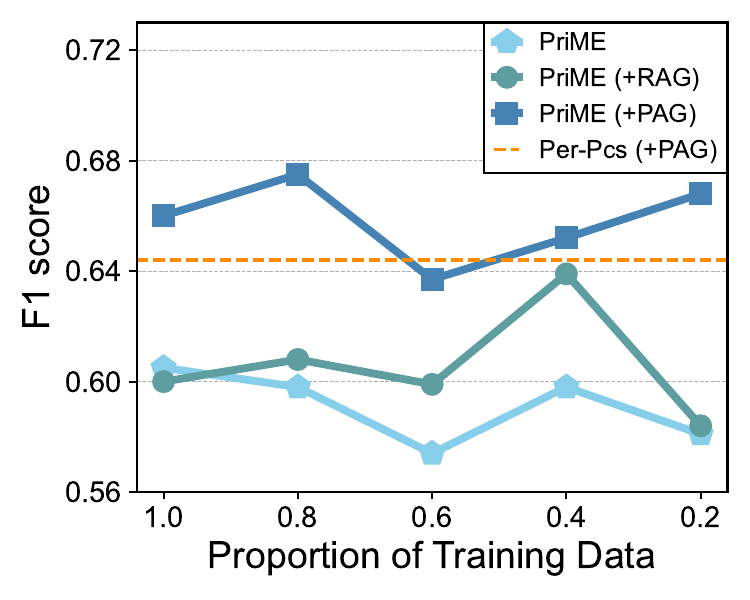}
        \caption{LaMP-1}
    \end{subfigure}
    \hfill \\
    \begin{subfigure}[t]{0.48\textwidth}
        \centering
        \includegraphics[width=0.49\linewidth]{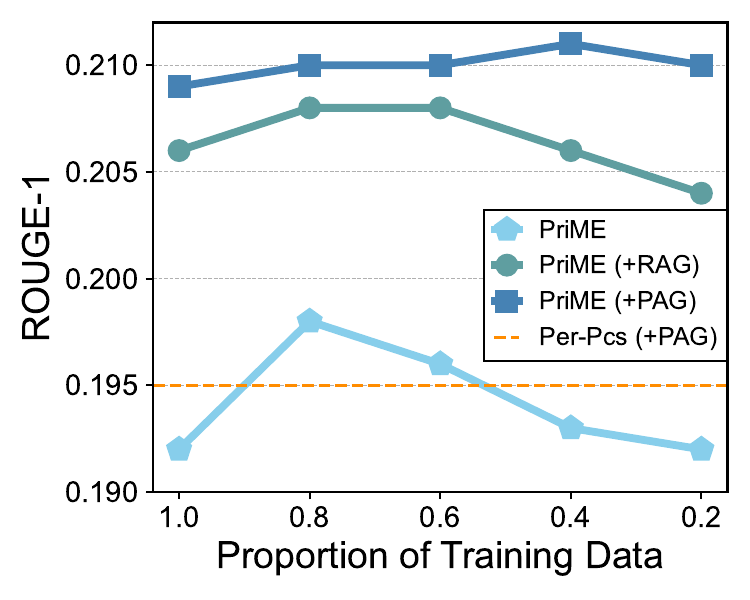}
        \includegraphics[width=0.49\linewidth]{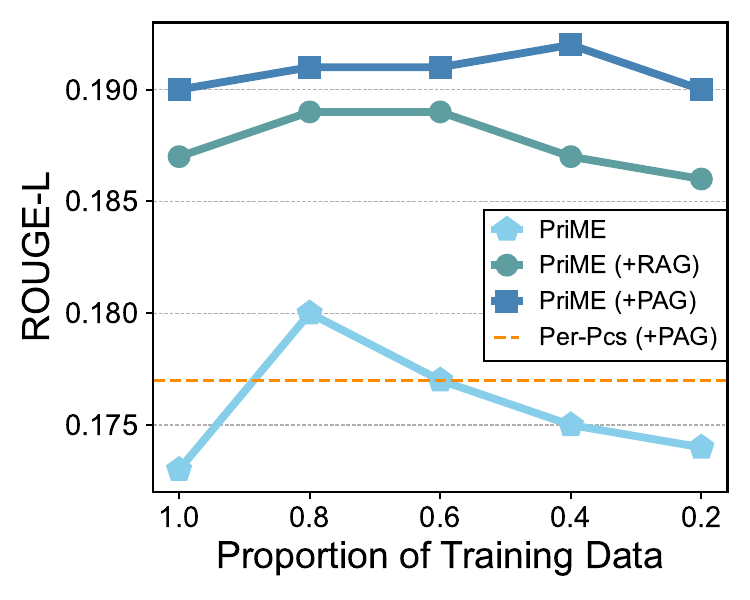}
        \caption{LaMP-4}
    \end{subfigure}
    \caption{\textbf{Effects of subsampling data.} {\sname} outperforms Per-Pcs, which uses the full dataset, often with just 20\% of the training data.}
    \vspace{-0.10in}
    \label{fig:train-data-sample-add}
\end{figure}

In Section~\ref{ss:ablations}, we show that training data can be reduced by 40\% while maintaining performance comparable to or better than Per-Pcs on LaMP-2.
As shown in Figure~\ref{fig:train-data-sample-add}, the amount of training data can be significantly reduced also on LaMP-1 and LaMP-4, often down to 20\%, while still achieving better test performance than Per-Pcs.

\subsection{Comparison of optimization algorithms}
\label{ss:appendix:add-results-opt}

\begin{figure}[t]
    \centering
    \begin{subfigure}[t]{0.48\textwidth}
        \centering
        \includegraphics[width=0.49\linewidth]{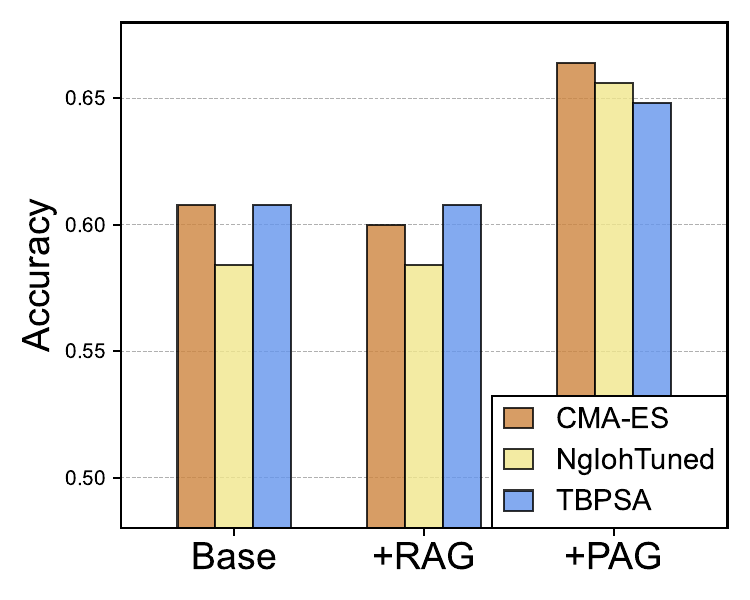}
        \includegraphics[width=0.49\linewidth]{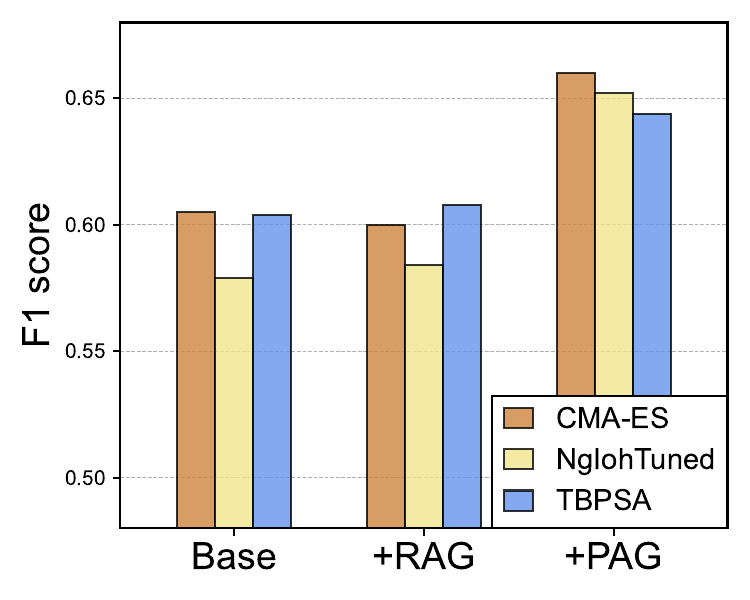}
        \caption{LaMP-1}
    \end{subfigure}
    \hfill \\
    \begin{subfigure}[t]{0.48\textwidth}
        \centering
        \includegraphics[width=0.49\linewidth]{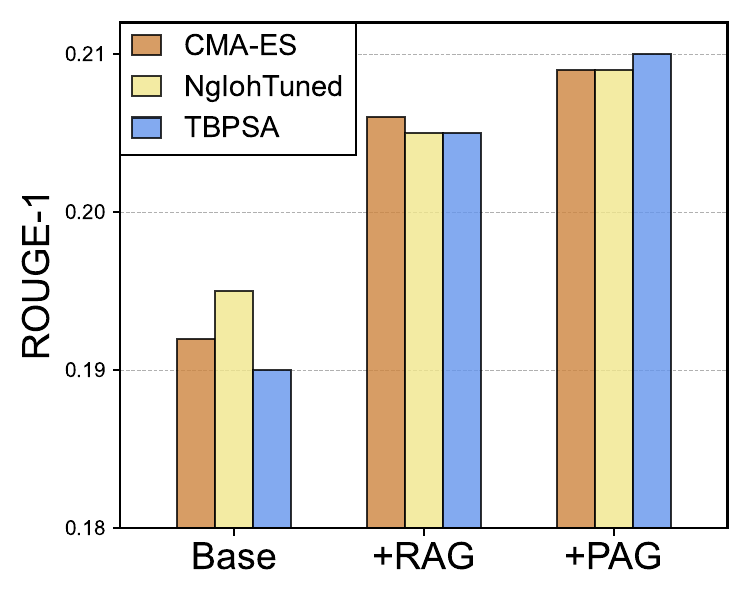}
        \includegraphics[width=0.49\linewidth]{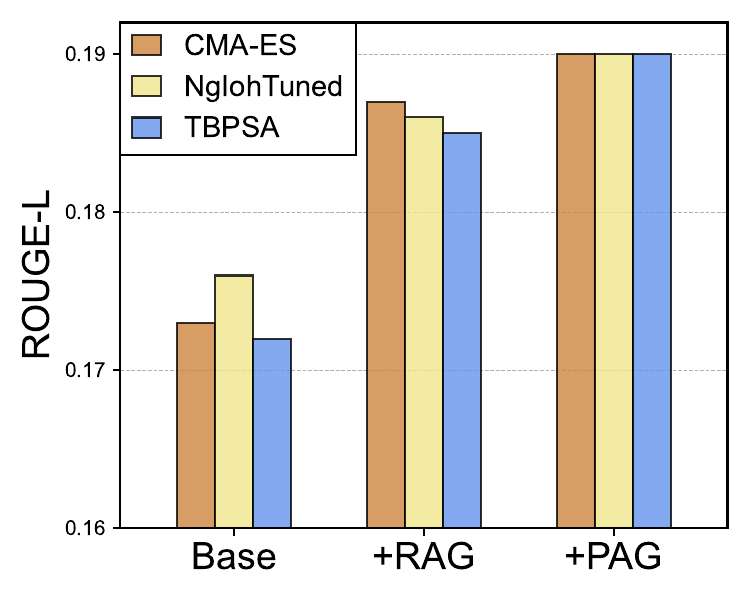}
        \caption{LaMP-4}
    \end{subfigure}
    \caption{\textbf{Comparison of optimizers.} {\sname} achieves competitive results across a range of gradient-free optimization algorithms.}
    \vspace{-0.10in}
    \label{fig:optimizers-add}
\end{figure}

As illustrated in Figure~\ref{fig:optimizers-add}, {\sname} consistently achieves competitive results across a range of gradient-free optimization algorithms on LaMP-1 and LaMP-4.
Similar to LaMP-2, more general-purpose optimizers, such as CMA-ES and NgIohTuned, demonstrate effectiveness with {\sname}.

\begin{table*}[t]
    \centering
    \footnotesize  %
    \setlength{\tabcolsep}{5pt}
    \begin{tabular}{ll}
        \toprule[1pt]
        \textbf{Source} & \textbf{Output} \\
        \midrule[0.75pt]
        Per-Pcs & 10 Living Room Decor Ideas That Will Make You Want To Throw A Party (PHOTOS) \\
        $\alpha = 0.0$ & 10 Living Room Decor Ideas That Will Make Your Home The Perfect Place To Entertain (PHOTOS) \\
        $\alpha = 2.0$ & 10 Ways to Make Your Living Room Look More Expensive \\
        Top sim. & 10 Living Room Decor Ideas That Will Make Your Home The Perfect Place To Entertain (PHOTOS) \\
        \midrule[0.50pt]
        Per-Pcs & 10 Things You Didn't Know About Presidential Vacations \\
        $\alpha = 0.0$ & The Most Expensive Vacations In The World \\
        $\alpha = 2.0$ & 10 Vacations You Can't Take \\
        Top sim. & World Leaders' Vacations (PHOTOS) \\
        \midrule[0.50pt]
        Per-Pcs & The Middle Class Is Under Attack \\
        $\alpha = 0.0$ & The Middle Class Is Dying. Here's How to Save It \\
        $\alpha = 2.0$ & The Trickle-Down Economy Is A Failure \\
        Top sim. & The Middle Class Is Dying. Here's How to Save It \\
        \bottomrule[1pt]
    \end{tabular}
    \vspace{-0.05in}
    \caption{\textbf{Qualitative comparisons.} Examples comparing Per-Pcs, {\sname} with varying $\alpha$, and the top similar user.}
    \label{table:app-qual}
\end{table*}

\subsection{Prompt sensitivity}
\label{ss:appendix:prompt-sens}

We conduct additional experiments to evaluate the sensitivity of prompt-based methods to (1) repeated decoding and (2) varied input queries.
For the repeated decoding experiment, we report the mean and standard error over three independent runs.
For the varied input queries experiment, we rephrase the original user profiles using GPT-4o-mini and evaluate how performance changes compared to the original profiles, using the same random seed for decoding.

\begin{table}[t]
    \centering
    \tiny  %
    \setlength{\tabcolsep}{5pt}
    \begin{tabular}{llccc}
        \toprule[1pt]
        \multirow{2.5}{*}{\textbf{Task}} & \multirow{2.5}{*}{\textbf{Metric}} & \multicolumn{1}{c}{\textbf{RAG}} & \multicolumn{2}{c}{\textbf{PAG}} \\ 
        \cmidrule(lr){3-3} \cmidrule(lr){4-5}
         & & $k=1$ & $k=0$ & $k=1$ \\
        \midrule[0.75pt]
        {\tiny LaMP-2} & Acc $\uparrow$ & ${0.412 \pm 0.014}$ & ${0.439 \pm 0.027}$ & ${0.467 \pm 0.020}$ \\
         & F1 $\uparrow$ & ${0.336 \pm 0.020}$ & ${0.340 \pm 0.019}$ & ${0.379 \pm 0.020}$ \\
        \midrule[0.50pt]
        {\tiny LaMP-4} & R-1 $\uparrow$ & ${0.202 \pm 0.001}$ & ${0.196 \pm 0.002}$ & ${0.207 \pm 0.001}$ \\
         & R-L $\uparrow$ & ${0.184 \pm 0.001}$ & ${0.177 \pm 0.001}$ & ${0.188 \pm 0.000}$ \\
        \bottomrule[1pt]
    \end{tabular}
    \vspace{-0.05in}
    \caption{\textbf{Variance from repeated decoding.} Performance is relatively consistent across repeated decoding.}
    \label{table:app-rep-dec}
\end{table}

\begin{table}[t]
    \centering
    \footnotesize  %
    \setlength{\tabcolsep}{5pt}
    \begin{tabular}{llcccc}
        \toprule[1pt]
        \multirow{2.5}{*}{\textbf{Task}} & \multirow{2.5}{*}{\textbf{Metric}} & \multicolumn{2}{c}{\textbf{Original}} & \multicolumn{2}{c}{\textbf{GPT}} \\ 
        \cmidrule(lr){3-4} \cmidrule(lr){5-6}
         & & $k=0$ & $k=1$ & $k=0$ & $k=1$ \\
        \midrule[0.75pt]
        {LaMP-2} & Acc $\uparrow$ & 0.470 & 0.490 & 0.419 & 0.441 \\
         & F1 $\uparrow$ & 0.361 & 0.402 & 0.316 & 0.354 \\
        \midrule[0.50pt]
        {LaMP-4} & R-1 $\uparrow$ & 0.194 & 0.207 & 0.194 & 0.188 \\
         & R-L $\uparrow$ & 0.176 & 0.206 & 0.175 & 0.187 \\
        \bottomrule[1pt]
    \end{tabular}
    \vspace{-0.05in}
    \caption{\textbf{PAG with different user profiles.} How user profiles are phrased in text can have a notable impact on PAG performance.}
    \label{table:app-pag-gpt}
\end{table}

Table~\ref{table:app-rep-dec} summarizes the results of the repeated decoding experiment on LaMP-2 and LaMP-4.
Overall, we observe low variance in performance across random seeds, suggesting that sampling has limited effect on the results.
Table~\ref{table:app-pag-gpt} presents results from the varied input queries experiment, where user profiles are rephrased using GPT-4o-mini.
In LaMP-2 (multi-class classification), profiles typically mention users' most frequently labeled movie genres (e.g., action, comedy), while in LaMP-4 (generation), they describe general writing styles.
On LaMP-4, performance with rephrased profiles is similar to the original, with a slight drop in the $k=1$ case.
In contrast, LaMP-2 shows a more noticeable decline.
Our analysis reveals that GPT often rephrased class labels such as ``sci-fi'' to ``science fiction'', leading to predictions that do not fully match the expected label set.
These findings suggest that prompt-based methods can be sensitive to how user profiles are phrased, particularly in classification tasks.

\subsection{Qualitative examples}
\label{ss:appendix:qual-examples}

To better illustrate qualitative differences among methods, we include in Table~\ref{table:app-qual} several example outputs from the benchmark.
Specifically, we compare outputs from Per-Pcs, {\sname} with $\alpha = 0.0$, {\sname} with $\alpha = 2.0$, and the output generated with the top similar user's module.
As shown by these examples, {\sname} tends to produce outputs more similar to those of the most similar user and, with a greater similarity penalty, generates more (often substantially) rephrased outputs.

\section{Prompt Templates}
\label{s:appendix:prompts}

For a fair comparison with the baseline results reported in \citet{tan2024personalized}, we adopt the same prompt templates and reproduce them below.
Table~\ref{table:prompt-profile} summarizes the prompt templates used to generate user profiles for each benchmark task, while Table~\ref{table:prompt-p13n} shows the prompt templates used for the actual personalization tasks.
Note that the prompts in Table~\ref{table:prompt-p13n} are prepended with the user profile, user history, or both, depending on whether RAG, PAG, or both are used.

\begin{table*}[ht]
    \centering
    \footnotesize  %
    \setlength{\tabcolsep}{5pt}
    \begin{tabular}[b]{p{0.11\linewidth}p{0.84\linewidth}}
        \toprule[1pt]
        \textbf{Task} & \textbf{Prompt Template} \\
        \midrule[0.75pt]
        \textsc{LaMP-1} & Write a summary, in English, of the research interests and topics of a researcher who has published the following papers. Only generate the summary, no other text. User History: \texttt{\{USER HISTORY\}} Answer: \\
        \midrule
        \textsc{LaMP-2} & Look at the following past movies this user has watched and determine the most popular tag they labeled. Answer in the following form: most popular tag: \texttt{<}tag\texttt{>}. User History: \texttt{\{USER HISTORY\}} Answer: \\
        \midrule
        \textsc{LaMP-3} & Based on this user's past reviews, what are the most common scores they give for positive and negative reviews? Answer in the following form: most common positive score: \texttt{<}most common positive score\texttt{>}, most common negative score: \texttt{<}most common negative score\texttt{>}. User History: \texttt{\{USER HISTORY\}} Answer: \\
        \midrule
        \textsc{LaMP-4} & Given this author's previous articles, try to describe a template for their headlines. I want to be able to accurately predict the headline given one of their articles. Be specific about their style and wording, don't tell me anything generic. User History: \texttt{\{USER HISTORY\}} Answer: \\
        \midrule
        \textsc{LaMP-5} & Given this author's previous publications, try to describe a template for their titles. I want to be able to accurately predict the title of one of the papers from the abstract. Only generate the template description, nothing else. User History: \texttt{\{USER HISTORY\}} Answer: \\
        \midrule
        \textsc{LaMP-7} & Given this person's previous tweets, try to describe a template for their tweets. I want to take a generic sentence and rephrase it to sound like one of their tweets, with the same style/punctuation/capitalization/wording/tone/etc. as them. Only give me the template description, nothing else. User History: \texttt{\{USER HISTORY\}} Answer: \\
        \bottomrule[1pt]
    \end{tabular}
    \vspace{-0.05in}
    \caption{\textbf{Profile generation prompts.} Prompt templates used for profile generation.}
    \label{table:prompt-profile}
    \vspace{-0.10in}
\end{table*}

\begin{table*}[ht]
    \centering
    \footnotesize  %
    \setlength{\tabcolsep}{5pt}
    \begin{tabular}[b]{p{0.11\linewidth}p{0.84\linewidth}}
        \toprule[1pt]
        \textbf{Task} & \textbf{Prompt Template} \\
        \midrule[0.75pt]
        \textsc{LaMP-1} & \#\#\# User Instruction: \\
        & Identify the most relevant reference for the listed publication by the researcher. Select the reference paper that is most closely related to the researcher's work. Please respond with only the number that corresponds to the reference. Paper Title: \texttt{\{QUERY PAPER TITLE\}} Reference: [1] - \texttt{\{OPTION1\}} [2] - \texttt{\{OPTION2\}} Answer: \\
        \midrule
        \textsc{LaMP-2} & \#\#\# User Instruction: \\
        & Which tag does this movie relate to among the following tags? Just answer with the tag name without further explanation. tags: [sci-fi, based on a book, comedy, action, twist ending, dystopia, dark comedy, classic, psychology, fantasy, romance, thought-provoking, social commentary, violence, true story] Description: \texttt{\{QUERY MOVIE DESCRIPTION\}} Tag: \\
        \midrule
        \textsc{LaMP-3} & \#\#\# User Instruction: \\
        & What is the score of the following review on a scale of 1 to 5? just answer with 1, 2, 3, 4, or 5 without further explanation. Review: \texttt{\{QUERY REVIEW\}} Score: \\
        \midrule
        \textsc{LaMP-4} & \#\#\# User Instruction: \\
        & Generate a headline for the following article. Article: \texttt{\{QUERY ARTICLE\}} Headline: \\
        \midrule
        \textsc{LaMP-5} & \#\#\# User Instruction: \\
        & Generate a title for the following abstract of a paper. Abstract: \texttt{\{QUERY ABSTRACT\}} Title: \\
        \midrule
        \textsc{LaMP-7} & \#\#\# User Instruction: \\
        & Paraphrase the following text into tweet without any explanation before or after it. Text: \texttt{\{QUERY TEXT\}} Tweet: \\
        \bottomrule[1pt]
    \end{tabular}
    \vspace{-0.05in}
    \caption{\textbf{Personalization prompts.} Prompt templates used for personalization.}
    \label{table:prompt-p13n}
    \vspace{-0.10in}
\end{table*}

\end{document}